\documentclass[journal]{IEEEtran}
\ifCLASSINFOpdf
   \usepackage[pdftex]{graphicx}
\else
\fi
\usepackage{amsmath}

\usepackage{amssymb}
\usepackage{mathtools}
\usepackage{booktabs}
\newcommand*{\ADDAFFILIATION}{}%
\usepackage{todonotes}
\usepackage{soul}

\DeclarePairedDelimiter\abs{\lvert}{\rvert}%
\DeclarePairedDelimiter\norm{\lVert}{\rVert}

\let\oldabs\abs
\def\abs{\@ifstar{\oldabs}{\oldabs*}}
\let\oldnorm\norm
\def\norm{\@ifstar{\oldnorm}{\oldnorm*}}

\begin{document}
%
\title{Improving Place Recognition Using Dynamic Object Detection}
%
%
%

\author{J. Pablo Mu\~{n}oz,~\IEEEmembership{Member,~IEEE,}
        and Scott Dexter
\ifdefined\ADDAFFILIATION
\thanks{J. Pablo Mu\~{n}oz is with Intel Labs, Intel Corporation, Santa Clara, CA, 95054 USA, e-mail: pablo.munoz@intel.com. \cite{MyThesis}.} 
\thanks{Scott Dexter is Professor of Computer Science at Alma College, Alma, MI 48801 USA, email: dextersd@alma.edu}
\thanks{Manuscript received June 11, 2020
}
\fi 
}

\maketitle

\begin{abstract}

We present a novel approach to place recognition well-suited to environments with many dynamic objects---objects that may or may not be present in an agent's subsequent visits. By incorporating an object-detecting preprocessing step, our approach yields high-quality place representations that incorporate object information. Not only does this result in significantly improved place recognition in dynamic environments, it also significantly reduces memory/storage requirements, which may increase the effectiveness of mobile agents with limited resources. 


\end{abstract}


%
\IEEEpeerreviewmaketitle

\section{Introduction}
%
%
%
%

\IEEEPARstart{A}{ppearance-based} place recognition is a crucial component of mapping, localization and navigation applications, which assist agents in their exploration of indoor and outdoor environments. By recognizing places, these agents can better plan their paths to a desired destination and/or correct errors when performing \emph{Simultaneous Localization and Mapping (SLAM)}. The importance of accurate and rapid visual place recognition is even more critical in situations where agents cannot rely on \emph{Global Positioning System (GPS)} or other technologies to confirm that they are revisiting a place, such as in indoor environments. 

Image-based approaches have proven to be robust methods for recognizing places \cite{Williams2009}. When agents use appearance-based place recognition, they attempt to infer their location from matching information about their current environment, gathered by their visual sensors, with a database of information about previously-visited locations. State-of-the-art devices that use sophisticated methods for appearance-based place recognition have shown outstanding performance in mapping and localization tasks \cite{LeeTango}. Researchers have exploited the capabilities of these devices in a variety of applications, including indoor navigation \cite{Munoz2016} \cite{munozCyber2017}.

Indoor and outdoor places alike are usually populated with dynamic objects, that is, objects that are not guaranteed to be present or in the same location in future observations of the place. Some of these objects may be in motion (such as a car driving by); others may be motionless (such as a parked car) but nonetheless temporary. A significant presence of these dynamic objects can cause traditional appearance-based place recognition algorithms to fail. In this article, we present a novel approach that augments traditional image-based place representation schemes with high-level visual information about dynamic objects, both moving and motionless. 

Our approach produces the following contributions: 
\begin{itemize}
\item improvement in the accuracy of place recognition in environments populated by dynamic objects;
\item  reduction in the time required to match two places; 
\item reduction in the size of the original representation used by ``flexible'' place recognition algorithms; and
\item reduction in the size of the database of places visited by an agent.
\end{itemize}

In addition, we define two related concepts:
\begin{itemize}
\item \emph{validity} of a place representation based on the presence of dynamic objects. We describe how this notion of a \emph{valid} place representation can be used to make efficiency improvements to traditional place recognition algorithms, and to measure the quality of an agent's observation; and
\item \emph{rigid} and \emph{flexible} existing place recognition techniques, a classification that depends on the malleability of their place representation schema.
\end{itemize}

The remainder of this article is organized as follows: Section \ref{relWork} discusses related work in appearance-based place recognition, object classification, and localization. Section \ref{approach} describes the proposed method to improve place representations. Section \ref{bbow} explains how the proposed method can be incorporated in state-of-the-art place recognition algorithms.  Section \ref{evaluation} presents an evaluation of the proposed approach. 

\section{Related Work}\label{relWork}

\subsection{Appearance-based Place Recognition}

Appearance-based place recognition approaches have substantially improved their effectiveness in the past few years, but there is still room for improvement. Early approaches were only capable of deciding whether an agent was visiting a particular room based on multiple images taken from multiple different viewpoints \cite{Ulrich2000}. More recent approaches are capable of localizing an agent with great accuracy based on a single image that is associated with a pose of the agent, e.g., \cite{Galvez-Lopez2012}
\cite{Cummins2011}
\cite{Sunderhauf2011}
\cite{Milford2012SeqSLAM}
\cite{Johns2013}\cite{Stumm2013}\cite{Pepperell2014}\cite{Arroyo2014}\cite{FastSeqSlam17}. These latter approaches use sophisticated human-crafted feature detectors and descriptors to produce robust place representations. Several feature detectors and binary descriptors, such as Learned Arrangements of Three Patch Codes (LATCH) \cite{DBLP:journals/corr/LeviH15}, produce compact and precise representations in a fraction of the time required by traditional approaches like \emph{Scale Invariant Feature Transform (SIFT)} \cite{Lowe1999} \cite{Lowe2004} and \emph{Speeded-Up Robust Features (SURF)} \cite{Bay:2008:SRF:1370312.1370556}. 
A breakthrough in local feature detection occurred with the development of \emph{Features from Accelerated Segment Test (FAST)} \cite{rosten_2006_machine}, a corner detector that incorporated the \emph{Univalue Segment Assimilating Nucleus (USAN)} principle \cite{Smith97} and machine learning techniques. Improvements to the \emph{FAST} detector produced \emph{Adaptive and Generic Corner Detection Based on the Accelerated Segment Test (AGAST)} \cite{mair2010_agast}, which uses a combination of generic decision trees instead of the environment-specific decision trees of the original \emph{FAST} algorithm.

Along with these successful feature detection and description techniques, \emph{Bags of Visual Words}  \cite{Sivic:2003:VGT:946247.946751} \cite{Li:2005:BHM:1068508.1069129} allows us to use feature descriptions as the basis for efficient image matching. By quantizing feature descriptors into  ``visual words'' using a distance metric, an image can be represented as a vector---the ``bag of visual words''---that collects the visual words in the image. Matching images then becomes a problem of finding images that have the most similar arrangement of visual words. Several improvements to this approach have been proposed throughout the years, with the \emph{vocabulary tree} being among the most successful \cite{Nister:2006:SRV:1153171.1153548}. 
\emph{FABMAP}, a turning point in place recognition frameworks, used bags of words to perform place recognition by modeling the correlation of visual words in an agent's observation \cite{Cummins2011}. Kejriwal et al. \cite{Kejriwal2016} proposed the use of an additional vocabulary of word pairs that has proven to be effective in dealing with the problem of \emph{perceptual aliasing}. 

More recently, the advent of binary descriptors made it easier to implement real-time place recognition applications, since these descriptors require orders of magnitude less construction time than approaches like \emph{SIFT} and \emph{SURF}. The \emph{BRIEF-Gist} \cite{Sunderhauf2011} approach to place recognition proved that using a very simple representation composed of a very small number of \emph{Binary Robust Independent Elementary Features (BRIEF)} \cite{Calonder2010} descriptors could yield performance levels competitive with more sophisticated approaches like \emph{FABMAP}. Later, the \emph{Bags of Binary Words}  \cite{Galvez-Lopez2012} approach showed how \emph{BRIEF} descriptors could be quantized into visual words to efficiently and accurately recognize places. The \emph{BRIEF} descriptor is not invariant to rotation and scale, but more sophisticated binary descriptors---such as \emph{Binary Robust Invariant Scalable Keypoints (BRISK)} \cite{Leutenegger2011}, \emph{Oriented FAST and Rotated BRIEF (ORB)} \cite{Rublee2011a},  and \emph{Fast Retina Keypoint (FREAK)} \cite{EPFL-CONF-175537}---which have greater robustness to changes in rotation, scale, viewpoint, and/or illumination, have supported advancements in place recognition systems. Some approaches use additional information to describe places. For instance, \emph{ABLE-S} adds depth information to the place representation in order to make it more robust \cite{Arroyo2014}. 

In the last decade, Deep Artificial Neural Networks have been successfully used to solve image classification problems \cite{Krizhevsky2012}\cite{Russak2015}. Appearance-based place recognition is closely related, and indeed, place recognition techniques incorporating Deep Learning techniques have shown promising results \cite{Chen2014}. For instance, approaches based on Convolutional Neural Networks (CNNs) can achieve real-time place recognition with great accuracy \cite{Sunderhauf2015}; Deep Learning techniques will continue to permeate place recognition in the near future. However, Deep Learning approaches require massive datasets for training that are not usually available for new environments in which place recognition will be performed; handcrafted feature detectors and descriptors are still fast and efficient solutions for place recognition systems. Some work is being done to improve place recognition within the CNN framework (e.g. \cite{Hou2017}), but in this article, we combine the two techniques, using ``traditional'' handcrafted feature detection and description augmented with Deep Learning-based detection of objects. We show the limitations of approaches that rely on handcrafted feature detection and description, especially in environments with a significant presence of dynamic objects, and we present effective solutions to overcome these limitations even in devices with limited resources.

 Furthermore, we show that by identifying and proposing solutions to the deficiencies of traditional approaches, we can also introduce useful notions, such as the validity of a place representation discussed in Section \ref{valid}.
 
\subsection{Object Detection and Recognition}

In this article, we use \emph{object detection} to improve the quality of low-level place representations, that is, those based on geometrical and/or topological information. Object detection and recognition can also be applied to the construction of \emph{semantic maps}, that is, maps that include additional high-level information about places \cite{KOSTAVELIS201745} \cite{KOSTAVELIS20131460}. 

The problem of identifying dynamic objects in an agent's visual observation is essentially a problem of \emph{image classification}. The goal of image classification is to assign a class to the whole image or a portion of it (in our case, the area that contains the detected object). Traditionally, researchers have used handcrafted features to recognize objects. Other work focuses on using  biologically-inspired techniques, such as the saliency maps of \cite{kostavelis2012}, to recognize and classify objects. Contemporary image classification techniques can produce highly accurate predictions. This success is primarily due to an embrace of Deep Learning, such as the techniques that showed drastically reduced image classification error rates in the \emph{ImageNet} competition \cite{Russak2015}. These error rates reached the single digits, which had never before happened with approaches relying on handcrafted feature detection and description. 

Deep Learning image classification techniques have been adapted to the problem of object detection. Among the most efficient and popular object detectors are unified, single-shot detectors, e.g., \emph{You Only Look Once (YOLOv3)} \cite{DBLP:journals/corr/RedmonDGF15} \cite{Redmon2016} 
or \emph{Single-shot Detector (SSD)} \cite{DBLP:journals/corr/LiuAESR15}, and two-stage detectors, e.g., \emph{Region-based CNN (R-CNN)}, \emph{Fast R-CNN} \cite{DBLP:journals/corr/Girshick15}, and \emph{Faster R-CNN} \cite{DBLP:journals/corr/RenHG015}. Below, we employ YOLO, because it provides real-time localization information (coordinates of the center of the object, width, height) and a confidence value of each detected dynamic object. Other methods may provide more accurate information about the detected dynamic objects, but they usually cannot be applied in real time.
\section{Combining Place Recognition and Dynamic Object Detection}\label{approach}

If an environment is densely populated by objects that do not have a permanent fixed position, agents may have great difficulty recognizing a previously-visited place. Traditional appearance-based place recognition approaches may extract features from the dynamic elements of a scene, essentially corrupting the representation of the place. If an agent returns to the same place, but a parked car has moved, or a bicyclist is traveling through, the agent may be unable to recognize the environment. 

Most place recognition algorithms use \emph{pose-based} representations, that is, places are represented by a multiset $pr$ of feature descriptors  $v_i, i = 1, 2,..., n$, generated from an agent's observation of a place from a particular pose.  
\begin{equation}\label{eq:pl}
pr = (v_{1}, v_{2}, ..., v_{n}).  
\end{equation} 
For instance, one version of the place recognition algorithm \emph{BRIEF-Gist} \cite{Sunderhauf2011}
represents a place with a single \emph{BRIEF} descriptor generated from a predetermined keypoint at the center of a downsampled image (the agent's observation).  That is, each $pr$ has size 1. The  \emph{FABMAP}
\cite{Cummins2011} algorithm, on the other hand, uses a vector of \emph{visual words} derived from an image representing the agent's observation. Each of these words are quantized descriptors that collectively represent a place; in this approach, the size of $pr$ may be in the hundreds.. Notably, in both techniques, the generated place representations may depend on pixels that are part of some dynamic object(s) in the scene.

In this article, we demonstrate an appearance-based place recognition approach that works by generating an ``ideal'' representation of a place, that is, one based only on those visual features that will be present and observable the next time an agent visits the place. That is, these ``ideal'' representations include no descriptors that describe, even in part, dynamic objects present in the agent's observation. How do dynamic objects affect descriptors in the place representation?  In general, a feature description procedure \emph{FD} takes a set of $n$ pixels, usually located around a selected or detected \emph{keypoint}, and produces a descriptor $v$ corresponding to the local feature at or near the keypoint. For instance, the ORB descriptor compares pairs of pixels in the vicinity of the keypoint to generate a binary descriptor. We define the \emph{extent} of  $v$ as the set of pixels $\{p_{1}, ..., p_{n}\}$, in the original image, $I$, that were used to generate $v$ (Equation \ref{eq:extent}). The descriptor may either depend directly on the pixels or it may transform them (e.g. by applying a filter to the original image). The extent may or may not include the feature keypoint.

\begin{equation}\label{eq:extent}
\text{extent}(v) = \{p_{i} \mid p_{i} \in I, 1\leq  i \leq n, FD(\{p_{1}, ..., p_{n}\}) \rightarrow v \}
\end{equation}We can classify each of the pixels in the original image as being part of either a dynamic or a static object. If the extent of the descriptor $v$ includes a pixel that belongs to a dynamic object, then we say that $v$ belongs to class \emph{DC}, the class of descriptors that are affected by dynamic objects. Otherwise, $v_i$ belongs to the class \emph{SC},  that is, the class of descriptors that are generated only from pixels that lie in static objects in the original image. Hence, $pr$ is the finite, pairwise disjoint multiset place representation that contains the union of \emph{DC} and \emph{SC}, 
\begin{equation}
pr = DC \cup SC. 
\end{equation}If a descriptor's extent contains just a few pixels from dynamic objects, it may not be effective to classify that descriptor in $DC$. We can relax the definition of $DC$ by defining a \emph{sensitivity threshold} indicating the proportion of pixels in the extent that belong to dynamic objects.  Thus, a descriptor $v$ is classified in \emph{DC} only when the proportion of pixels in the extent belonging to a dynamic object exceeds the sensitivity threshold.

In the following section, we use these ideas to classify popular place recognition algorithms based on their place representations. Then, we show how our proposed approach overcomes the limitations of traditional place recognition algorithms in environments highly populated by dynamic objects. Finally,  in \ref{detDynObj}, we discuss how to use Deep Learning-based object detectors and common properties in feature descriptors, e.g., they tend to be isotropic, to quickly estimate which descriptors belong to $DC$. 

\section{Incorporating Dynamic Objects into Place Recognition Algorithms
}\label{bbow}

Not every place recognition algorithm can be adapted to our approach. Depending on how place representations are constructed, there may be no way to eliminate the negative impact of dynamic objects---there may be no mechanism by which we can take into account information about dynamic objects in the agent's observation. For example,  both \emph{BRIEF-Gist} \cite{Sunderhauf2011} and \emph{ABLE-S} \cite{Arroyo2014} 
rely on a predetermined pattern of keypoints and sampled pixels. Because the underlying algorithm in each case depends on each of these descriptors with predetermined locations, we cannot remove any descriptors, even if we determine them to be in \emph{DC}. We classify place representation approaches as either \emph{rigid} or \emph{flexible}, depending on whether their representation scheme can be modified to  remove the impact of dynamic objects present in the place. Table \ref{tab:placeRepClass} gives our classification of a few popular place recognition approaches. 

\begin{table}
\begin{center}
\begin{tabular}{ |l|c| }
  \hline
  \multicolumn{2}{||c||}{Place Representation} \\
  \hline
  \hline
  \multicolumn{1}{|c|}{Approach} & \multicolumn{1}{|c|}{Representation} \\
  \hline
  FABMAP \cite{Cummins2011} & flexible \\
  \hline
  BRIEF-Gist \cite{Sunderhauf2011} & rigid \\
  \hline
  SeqSLAM \cite{Milford2012SeqSLAM} & rigid\\
  \hline
  Bags of Binary Words \cite{Galvez-Lopez2012} & flexible \\ 
  \hline
  Cooc-Map \cite{Johns2013} & flexible \\ 
  \hline
  COVISMAP \cite{Stumm2013} & flexible \\
  \hline
  SMART \cite{Pepperell2014} & rigid \\
  \hline  
  ABLE-S \cite{Arroyo2014} & rigid \\
  \hline
  Fast-SeqSLAM \cite{FastSeqSlam17} & rigid \\
  \hline
\end{tabular}
\end{center}
\caption{Examples of place recognition algorithms and our classification of their respective place representations.} \label{tab:placeRepClass}
\end{table}

\subsection{Bags of Binary Words}\label{BoW}

In the \emph{Bags of Binary Words} (BoBW) approach proposed by G\'{a}lvez-L\'{o}pez et al. \cite{Galvez-Lopez2012},  it is possible to modify the place representation to take into account the presence of dynamic objects. In \ref{evaluation},  we adapt this approach in order to evaluate the effectiveness of our technique. Below, we briefly describe BoBW.  This approach was the first to use binary descriptors with the \emph{Bag of Visual Words} paradigm. Initially, BoBW used \emph{BRIEF} descriptors, but other implementations  use \emph{ORB} descriptors \cite{Rublee2011a}, which have the added advantage of rotation invariance.   

In the BoBW paradigm, first, a vocabulary tree is built from the discretization of the binary descriptor space. The final structure, a hierarchical tree, allows for efficiently matching place representations (i.e. bags of visual words). By using binary descriptors and the Hamming distance, BoBW is capable of reducing the computation time required for matching bags of visual words by an order of magnitude compared to the time required by other popular approaches, e.g., \cite{Cummins2011} and \cite{Stumm2013}. 

BoBW uses an \emph{inverted index}, a common structure used in \emph{Bag of Visual Words} approaches, to quickly find images where a particular word is present. That is, if we have a collection of images $I_{t}$, each described by a ``bag of words'' \emph{bag($I_{t}$}, the inverted index allows us to ``look up'' an individual word and find all the images containing that word in their description.  G\'{a}lvez-L\'{o}pez et al. augment this index to include the weight of the word in the image, so the inverted index maps words to sets of pairs $w_{i}\rightarrow{\langle t, v_{t}^{i} \rangle}$. That is, if word $w_{i}$ is present in the bag of words describing  image $I_{t}$ and $v_{t}^{i}$ is the weight of the visual word $w_{i}$ in $I_{t}$, then the index entry for word $w_{i}$ is 
\begin{equation}\label{eq:invertedIndex}
i : \{<t, v_{t}^{i}> \mid w_{i} \in bag(I_{t})\}.
\end{equation}In addition to the \emph{inverted index}, G\'{a}lvez-L\'{o}pez et al. also introduce a \emph{direct index} to store a reference to the features extracted from the image. This index plays an important role when checking for geometrical consistency. Using this index, G\'{a}lvez-L\'{o}pez et al. can quickly access a subset of the features of the candidate image, and together with the features from the query image, they compute a fundamental matrix using \emph{Random Sample Consensus} (RANSAC)\cite{Fischler1981}. The direct index is used to avoid comparing all the features in the pair of images when verifying for geometrical consistency. Geometric verification can be disabled, per level, or exhaustive, i.e., using all detected features.

G\'{a}lvez-L\'{o}pez et al. use a $L_{1}$-score (Equation \ref{eq:score}) to measure the similarity between two binary bags of words, $\mathbf{v_{1}}$ and $\mathbf{v_{2}}$: 
\begin{equation}\label{eq:score}
s(\mathbf{v_{1}}, \mathbf{v_{2}}) = 1 - \frac{1}{2}\left|\frac{\mathbf{v_{1}}}{\
\lvert\mathbf{v_{1}}\rvert} - \frac{\mathbf{v_{2}}}{\lvert\mathbf{v_{2}}\rvert}\right|
\end{equation}
This score is a scaled version of the score proposed by Nister et al. in their seminal paper about creating hierarchical trees of words \cite{Nister:2006:SRV:1153171.1153548}. 

\subsection{Determining Whether a Descriptor is Affected by Dynamic Objects}\label{detDynObj} 

To determine whether a descriptor $v_{i} \in pr$ is a member of \emph{DC}, we need to identify the areas occupied by dynamic objects in the image. A fast object detector, e.g., \emph{YOLO}\cite{Redmon2016}, can be used to obtain the approximate area occupied by a dynamic object in real time. The object detector produces bounding boxes that roughly enclose the detected dynamic objects; with these boxes, we can find the descriptors that are  affected by a dynamic object above the sensitivity threshold. But in the case of some complex feature descriptors, measuring the proportion of a descriptor's extent that is based on dynamic objects may be very computationally expensive.

Alternatively, we can use heuristics that take advantage of common properties of feature descriptor algorithms. For example, many feature descriptor algorithms sample locations in an isotropic manner around the feature keypoint. Hence, one heuristic is that if the keypoint is located inside the bounding box of a dynamic object, we can conclude that at least 25\% of the extent of the descriptor is affected by dynamic objects. This is particularly useful if we set the sensitivity y threshold at $\frac{1}{4}\vert \text{extent}(v) \vert$---then we simply define \emph{DC} to be the class of all descriptors whose keypoints are inside a bounding box. Another heuristic works well for a sensitivity threshold of  $\frac{1}{2}\vert \text{extent}(v) \vert$: by using the distance $r$ from the keypoint of $v$ to the furthest sampled point in $\text{extent}(v)$, we can identify the keypoints inside a bounding box and more than $r$ pixels from each corner; these descriptors will be in \emph{DC} for sensitivity threshold $\frac{1}{2}\vert \text{extent}(v) \vert$.

\begin{figure}[!th]
\centering
   \includegraphics[width=0.6\columnwidth]{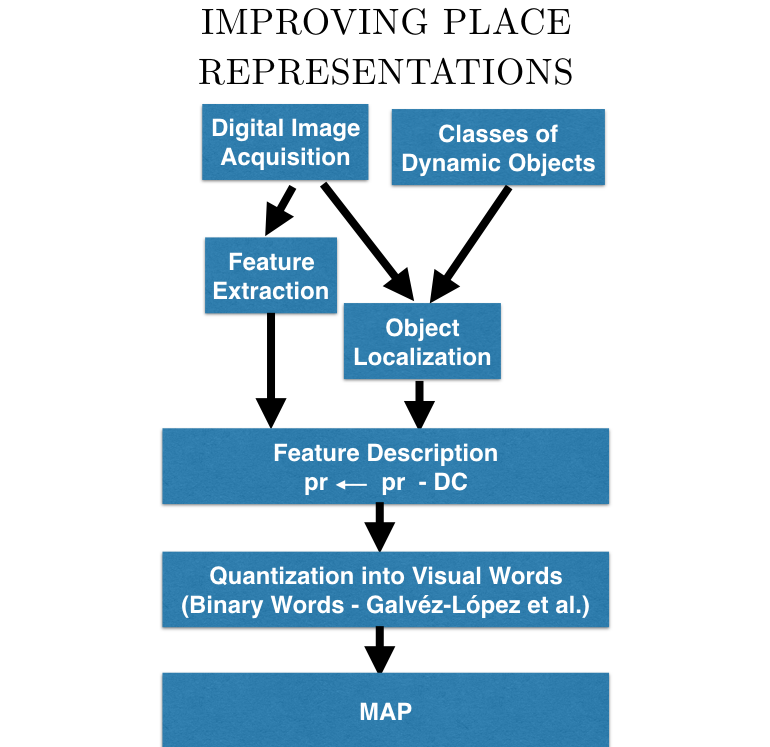}
   \caption{Diagram of how incorporating the proposed procedure improves a place representation by taking into account high-level information from dynamic objects.
   }~\label{fig:improvingPlaceR}
\end{figure}

Figure \ref{fig:improvingPlaceR} illustrates the method to improve a place representation based on dynamic object information. The procedure receives a list of dynamic objects of interest to be detected in the captured images. Using the information from the object detector, place representations are modified to reduce the impact of descriptors that are affected by dynamic objects. 

\subsection{Valid Place Representation and Efficiency Improvements}\label{valid} 
Two or more observations of the same place in the real world can result in several different place representations. One reason is that these images may contain dynamic objects, which may alter the representation of the place, resulting in alternative representations. Ideally, once an agent has captured a digital image of a place, the generated representation should be robust enough to allow the agent to match it with a representation of a future observation of the place. Incorporating high-level information about dynamic objects when generating a place representation allows us to define the concept of a \emph{valid} place representation. 

An arbitrary place representation, $pr_i$ in the set of place representations of an environment and generated at step $i$, is valid if it contains a number of descriptors from the class \emph{SC} that is above a threshold, \emph{placeThreshold} (Equation \ref{valideq}). That is, all of these descriptors in the place representation have an extent below the \emph{sensitivity} threshold defined in Section \ref{approach}. $prs_i$ is the optimized place representation that contains only descriptors from the class \emph{SC}, unaffected by dynamic objects.

\begin{equation}
\text{isValid}(pr_{i}) = 
\begin{cases}
 \text{ true} & \text{if}\  \vert prs_i \vert > \text{placeThreshold}, \\ & prs_{i} = pr_{i} - DC, \\
  & \text{i.e.}, \forall v(v \in prs_i \rightarrow v \in SC), \\
  \text{false} & \text{otherwise}.\ 
\end{cases}\label{valideq}
\end{equation}

We can use this idea to implement at least two kinds of efficiency improvements, assuming we have a \emph{flexible} place recognition systems. One kind of improvement occurs at the level of \emph{place}. First, an agent might decide not to store invalid place representations, resulting in reduced storage requirements. If invalid place representations \emph{are} stored, an agent can avoid the costly procedure of attempting to match a place that has no valid representation. At the level of the \emph{place representation}, we can reduce the size of place representations by storing only descriptors in \emph{SC}. These reductions accumulate to yield a significantly smaller database, which is crucial for exploration of large environments and/or devices with limited storage. Additionally, the computation time required to match two images (i.e. to recognize an already-visited place) will be much reduced for smaller place representations.

Traditional place recognition algorithms do not  discriminate between observations. They attempt to find a match in the database for each new observation, even when these observations produce a place representation with a small number of descriptors. What is worse is that, as we have mentioned in this article, traditional place recognition algorithms do not take into account that despite the number of descriptors in a place representation, some of those descriptors may be generated from dynamic objects, hence misrepresenting the place in question. Having bad quality place representations in the database increases its size and makes the system inefficient. To the best of our knowledge, we are the first to introduce the concept of a valid place representation, and use it to discriminate observations based on the quality of the detected features.
\section{Evaluation}~\label{evaluation}

\subsection{Experimental Configuration}  
The proposed approach was evaluated using a Dell Precision 5510 workstation running Ubuntu 16.04LTS with 8GiB of RAM, an Intel Core i7-6700HQ processor, and an Nvidia Quadro M1000M GPU. We used two datasets in the evaluation, one with synthetic images (Synthia dataset \cite{RosCVPR16}), and the other containing real-world images (M\'{a}laga dataset \cite{blanco2013mlgdataset}). 

\begin{figure}[!th]
\centering
   \includegraphics[width=0.8\columnwidth]{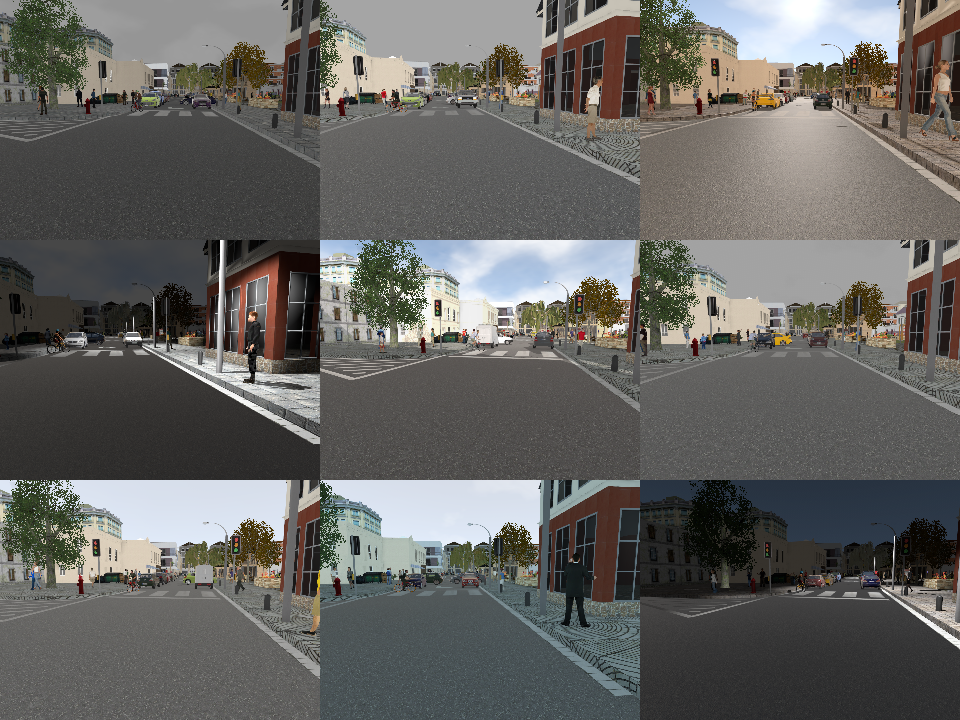}
   \caption{Collage of images from the Synthia dataset corresponding to the same location with different illumination, textures, and dynamic objects.}~\label{fig:synthiaCollage}
\end{figure}

We used the \emph{SYNTHIA-RAND-CVPR16} subset of the \emph{Synthia} dataset, which is a collection of photo-realistic frames taken every 10 meters as an agent moves in a virtual city. For each position, several frames are randomly generated using different configurations (illumination and textures), including a variation in the presence of different classes of dynamic objects. Figure \ref{fig:synthiaCollage} shows an example of the frames that correspond to one particular virtual location. In our evaluation with this dataset, we used the images from  the front camera, which is a subset of 4,485 images. In the case of the real world images from the \emph{M\'{a}laga} dataset, we used 17,300 images of subset \#10 that were captured at 20 frames per second in 865 seconds by a vehicle moving through the Spanish city of M\'{a}laga. For both datasets, we configured our system for high-level detection of the following dynamic objects: cars, trucks, motorcycles, bicycles (either moving or parked), and people (either standing in the sidewalks or walking). 

We used the vocabulary of binary words created from ORB descriptors \cite{Mur-Artal2015}, and the implementation of BoBW called  \emph{DBoW2}, by \cite{Galvez-Lopez2012}. We tested our approach with several configurations of the object detection, place representation, and place recognition parameters; see Table \ref{table:parameters}. For the configurations that required geometric verification, we used the default values in the \emph{DBoW2} library. 

For the identification of areas occupied by dynamic objects in an image we used the You Look Only Once (YOLO) object detection algorithm \cite{Redmon2016}, which works on square (1:1 aspect ratio) images in the RGB color space. Because the images in our dataset are not square, we cropped equal amounts from each side of the images. We then applied YOLO with weights determined by pre-training with the COCO dataset \cite{DBLP:journals/corr/LinMBHPRDZ14} to the squared RGB images. 

\begin{table}[!th] 
\begin{center}
  \begin{tabular}{ | l | c | }
   	\hline
	Parameter & Values \\
	\hline    
    \hline
    ORB Keypoints & 300, 500, 1000, 1500, 2000\\
    \hline
    Geometric verification & Disabled, level 0, Exhaustive check
    \\ 
    \hline
    YOLO Confidence Threshold & 0.10, 0.20, 0.30, 0.40\\
    \hline
    Sensitivity Threshold & 25\%\\
    \hline
  \end{tabular}
\end{center}
\caption{Configuration parameters for the evaluation.}\label{table:parameters} 
\end{table}

\subsection{Problem Formulation}
\begin{figure}[!th]
  \centering
   \includegraphics[width=0.8\columnwidth]
{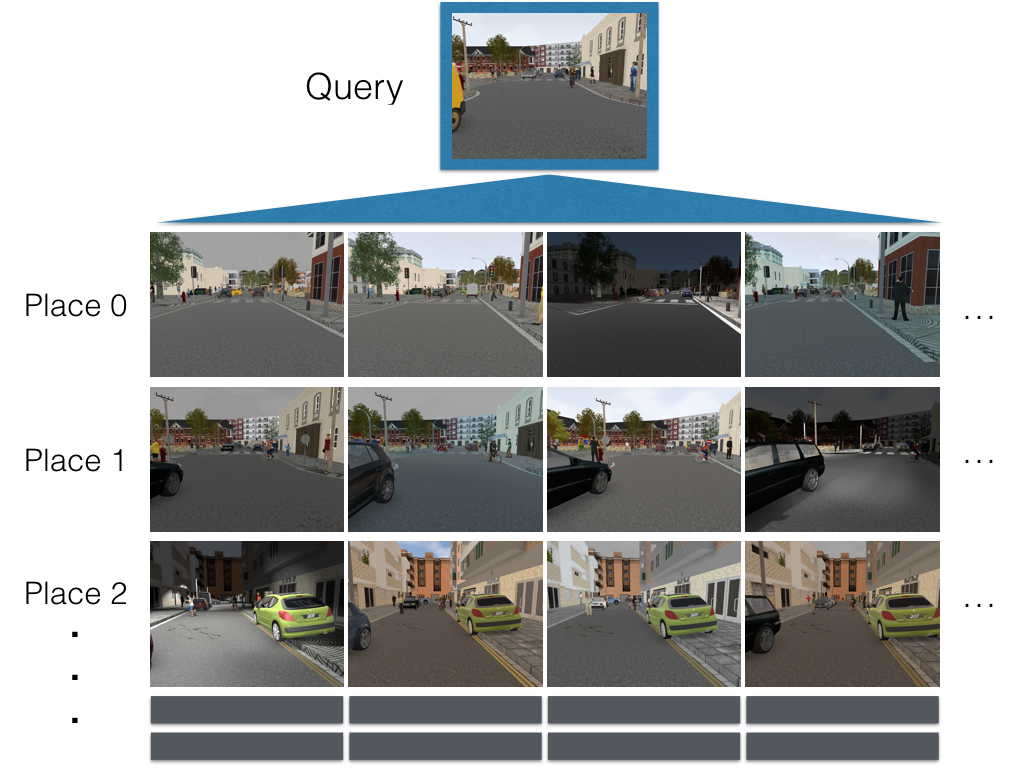}
   \caption{The agent has to identify other place representations associated with the place observed in the query image.}~\label{fig:synthiaQuery}
\end{figure}

In our evaluation, we focus on the scenario in which an agent has already captured observations of several configurations for each place. What occurs when the agent is given a new image of a place? Can the agent match this new image to one of the other representations of the the same place in the database? The problem is illustrated in Figure \ref{fig:synthiaQuery}. We compare the performance of the traditional Bag of Binary Words method with our extended version incorporating information about dynamic objects. 

\begin{figure*}[!th]
\centering
   \includegraphics[width=1\textwidth]{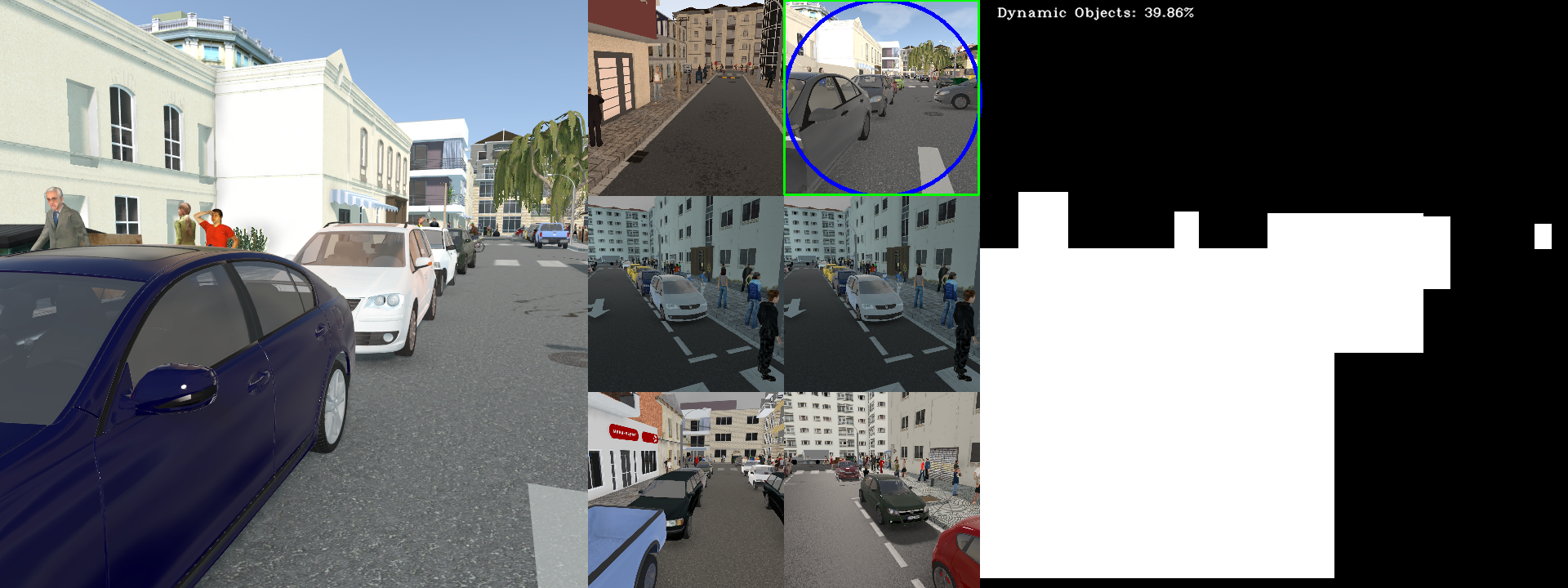}
   \caption{Place recognition on the Synthia dataset. On the left is the current observation. The first column in the middle shows the candidates found by the \emph{Bags of Binary Words} approach. The second column in the middle shows the candidates found by the extended approach, which incorporates knowledge about dynamic objects (the blue circle means that the candidate also passed geometric verification). On the right is the YOLO approximation of the space occupied by the dynamic objects in the image. The first candidate from our approach shows a correct prediction, even though the cars that are parked on the street are different from one observation to the next. The original approach fails to return a good match due to the presence of dynamic objects.}~\label{fig:synthiaExample}
\end{figure*}

\begin{table*}[!th] 
\centering
\resizebox{\textwidth}{!}{
\begin{tabular}{|r|l|l|r|r|r|r|r|r|r|r|r|r|r|r|}
\toprule
\multicolumn{2}{|c|}{Configuration} & 
\multicolumn{3}{|c|}{All Images} &             \multicolumn{3}{|c|}{Images with $>$ 10\% Dyn. Obj.} & 
\multicolumn{3}{|c|}{Images with $>$ 20\% Dyn. Obj.} &
\multicolumn{3}{|c|}{Images with $>$ 30\% Dyn. Obj.}\\
\hline
keys &  Geom & 
BoBW &  BoBW+DO &  + - &  
BoBW &  BoBW+DO &  + - &
BoBW &  BoBW+DO &  + - &
BoBW &  BoBW+DO &  + -\\
\midrule
	300 & NoGeom & 42.9 & 47.78 & \bfseries 11.38 & 39.94 & 48.51 & \bfseries 21.46 & 39.51 & 49.59 & \bfseries 25.51 & 38.56 & 50.9 & \bfseries 32 \\ \hline
	300 & Geo-0 & 0.76 & 0.42 & -44.12 & 0.2 & 0.08 & -60 & 0.16 & 0 & -100 & 0.51 & 0 & -100 \\ \hline
	300 & Geo-1 & 2.92 & 2.23 & -23.66 & 1.87 & 1.24 & -34.04 & 2.03 & 1.3 & -36 & 2.31 & 1.8 & -22.22 \\ \hline
	300 & Geo-2 & 7.22 & 7.98 & \bfseries 10.49 & 5.22 & 6.7 & \bfseries  28.24 & 4.72 & 6.1 & \bfseries 29.31 & 4.63 & 6.17 & \bfseries 33.33 \\ \hline
	300 & Geo-6 & 23.75 & 23.14 & -2.54 & 20.96 & 22.28 & \bfseries  6.27 & 19.76 & 20.65 & \bfseries 4.53 & 18.51 & 20.82 & \bfseries 12.5 \\ \hline
	500 & NoGeom & 54 & 58.39 & \bfseries 8.13 & 52.41 & 58.71 & \bfseries 12.02 & 51.63 & 57.24 & \bfseries 10.87 & 51.16 & 60.41 & \bfseries 18.09 \\ \hline
	500 & Geo-0 & 5.73 & 4.53 & -21.01 & 4.66 & 3.47 & -25.64 & 5.45 & 3.41 & -37.31 & 7.97 & 4.63 & -41.94 \\ \hline
	500 & Geo-1 & 15.18 & 13.76 & -9.4 & 13.31 & 12 & -9.88 & 13.25 & 11.54 & -12.88 & 15.17 & 12.85 & -15.25 \\ \hline
	500 & Geo-2 & 12.91 & 17.35 & \bfseries 34.37 & 11.28 & 17.54 & \bfseries 55.48 & 11.63 & 17.8 & \bfseries 53.15 & 12.08 & 21.34 & \bfseries 76.6 \\ \hline
	500 & Geo-6 & 42.5 & 43.9 & \bfseries 3.31 & 40.89 & 44.08 & \bfseries 7.8 & 40.49 & 42.85 & \bfseries 5.82 & 39.85 & 46.27 & \bfseries 16.13 \\ \hline
	1000 & NoGeom & 63.95 & 68.38 & \bfseries 6.94 & 62.93 & 68.23 & \bfseries 8.42 & 62.03 & 68.13 & \bfseries  9.83 & 61.44 & 67.61 & \bfseries 10.04 \\ \hline
	1000 & Geo-0 & 28.18 & 27.31 & -3.09 & 26.94 & 26.19 & -2.81 & 26.91 & 25.12 & -6.65 & 30.33 & 28.79 & -5.08 \\ \hline
	1000 & Geo-1 & 28.74 & 34.27 & \bfseries 19.24 & 27.3 & 34.28 &  \bfseries 25.55 & 28.29 & 34.88 & \bfseries 23.28 & 32.39 & 38.3 & \bfseries 18.25 \\ \hline
	1000 & Geo-2 & 14.4 & 22.83 & \bfseries 58.51 & 12.71 & 24.07 & \bfseries 89.34 & 13.33 & 26.42 & \bfseries  98.17 & 15.42 & 30.85 & \bfseries 100 \\ \hline
	1000 & Geo-6 & 61 & 64.93 & \bfseries 6.43 & 59.43 & 64.29 & \bfseries 8.18 & 58.37 & 64.47 & \bfseries 10.45 & 58.87 & 65.81 & \bfseries 11.79 \\ \hline
	1500 & NoGeom & 69.54 & 74.23 & \bfseries 6.73 & 68.63 & 74.33 & \bfseries 8.3 & 66.42 & 74.47 & \bfseries 12.12 & 64.27 & 75.58 & \bfseries 17.6 \\ \hline
	1500 & Geo-0 & 37.17 & 41.07 & \bfseries 10.5 & 35.55 & 40.97 & \bfseries 15.25 & 34.47 & 41.06 & \bfseries 19.1 & 34.45 & 43.44 & \bfseries 26.12 \\ \hline
	1500 & Geo-1 & 33 & 40.78 & \bfseries 23.58 & 30.65 & 41.89 & \bfseries 36.67 & 29.35 & 43.66 & \bfseries 48.75 & 29.56 & 46.53 & \bfseries 57.39 \\ \hline
	1500 & Geo-2 & 16.95 & 25.93 & \bfseries 53.03 & 14.03 & 27.46 & \bfseries 95.74 & 13.25 & 29.92 & \bfseries 125.77 & 13.11 & 34.7 & \bfseries 164.71 \\ \hline
	1500 & Geo-6 & 67.92 & 72.4 & \bfseries 6.6 & 67.04 & 72.82 & \bfseries 8.62 & 64.63 & 72.6 & \bfseries 12.33 & 62.21 & 72.24 & \bfseries 16.12 \\ \hline
	2000 & NoGeom & 73.04 & 76.74 & \bfseries 5.07 & 71.86 & 76.76 & \bfseries 6.82 & 71.95 & 77.32 & \bfseries 7.46 & 73.52 & 79.18 & \bfseries 7.69 \\ \hline
	2000 & Geo-0 & 42.81 & 48.41 & \bfseries 13.07 & 41.61 & 48.74 & \bfseries 17.15 & 41.87 & 50.49 & \bfseries 20.58 & 44.73 & 57.33 & \bfseries 28.16 \\ \hline
	2000 & Geo-1 & 35.05 & 43.55 & \bfseries 24.24 & 31.89 & 45.64 & \bfseries 43.12 & 31.79 & 49.51 & \bfseries 55.75 & 34.7 & 56.3 & \bfseries 62.22 \\ \hline
	2000 & Geo-2 & 24.64 & 30.7 & \bfseries 24.62 & 22.52 & 31.85 & \bfseries 41.42 & 21.63 & 36.83 & \bfseries 70.3 & 24.42 & 44.47 & \bfseries 82.11 \\ \hline
	2000 & Geo-6 & 72 & 75.3 & \bfseries 4.58 &  70.75 & 74.81 & \bfseries 5.75 & 70.98 & 75.45 & \bfseries 6.3 & 72.24 & 77.12 & \bfseries 6.76 \\ \hline
\end{tabular}
}
\caption{Place recognition accuracy: original BoBW) algorithm and our extended approach (BoBW+DO). Incorporating information about dynamic objects improves the recognition rate in all configurations in which the recognition rate is greater than about 30\%.}\label{table:synthiaSummaryTable}
\end{table*}

\begin{figure*}[!th]
\centering
   \includegraphics[width=1\textwidth]{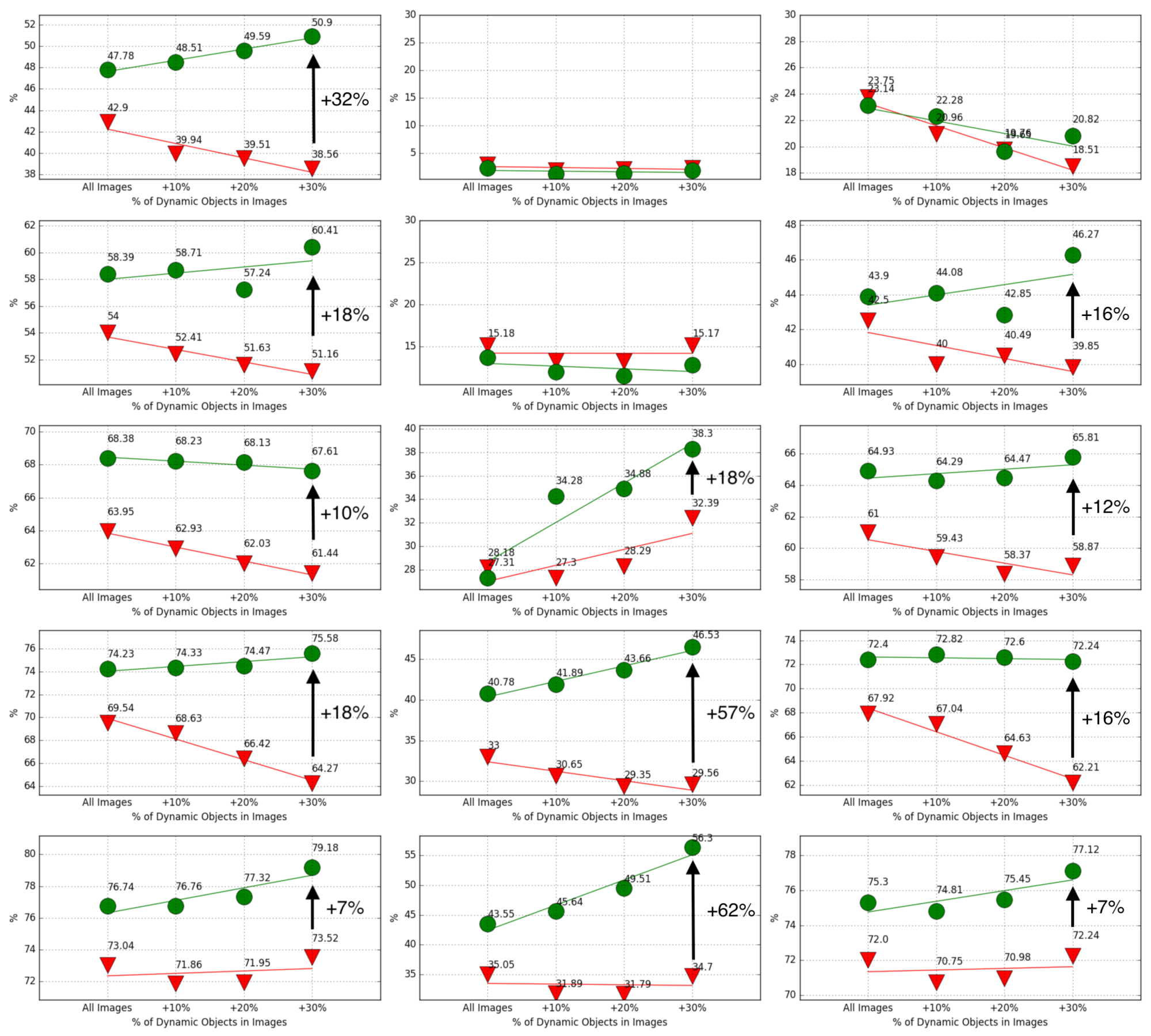}
   \caption{Percentage of correct place recognition in the Synthia dataset. Red triangles correspond to the  original Binary Bags of Words algorithm; green dots are the results when incorporating information from dynamic objects. Each row represents the approximate number of features extracted from each image (approximately 300, 500, 1000, 1500 and 2000), each column represents the degree of geometric verification used (no geometric verification, geometric verification at level 1, and exhaustive geometric verification). As the percentage of the area of the image that is covered by dynamic objects increases, the performance of our approach yields better place recognition
.}~\label{fig:synthiaSummary}
\end{figure*}

\subsection{Results}
Figure \ref{fig:synthiaExample} illustrates the difference in behavior between the original \emph{Bags of Binary Words} algorithm and our proposed enhancement. On the left is a picture of the current observation of the agent. The adjacent column of three images are candidate matches identified by BoBW; because of the presence of dynamic objects, none of these candidates are good matches. The next column of images are the candidates identified by our extended algorithm. The first candidate from our approach is a correct match, even though the cars that are parked on the street are different from one observation to the next (the blue circle indicates that our approach has also passed geometric verification). On the far right is the YOLO approximation of the dynamic objects detected in the observation. 

Table \ref{table:synthiaSummaryTable} 
shows a comparison of the results obtained by the original \emph{(BoBW)} approach and the proposed extended approach using dynamic objects to improve the place representation \emph{(BoBW + DO)}. This table shows how taking into account information about dynamic objects improves recognition results in all configurations in which the \emph{BoBW}-only recognition accuracy is more than about 30\%. When we further limit our analysis to those images with a minimum level of coverage by dynamic objects (10\%, 20\% and 30\%), our proposed approach performs much better than \emph{BoBW}-only approach as the percentage of dynamic objects in the images increases. The table shows only a subset of the results, with \emph{YOLO}'s confidence set to 0.20. Additional details are available in \cite{MyThesis}. 
Figure \ref{fig:synthiaSummary} shows that in most configurations, as the percentage of the area of the image that is covered by dynamic objects increases, the performance of our approach yields better place recognition than the \emph{Bags of Binary Words} approach without incorporating dynamic object detection. These improvements confirm the significance of our approach: incorporating high level information about dynamic objects improves the performance of existing place recognition algorithms in environments highly populated by dynamic objects. The place recognition accuracy improves significantly for images with a greater percentage of the area covered by dynamic objects. For instance, as shown in table \ref{table:synthiaSummaryTable}, when using 2000 ORB features, and geometric verification at level 1, the proposed approach yields a place recognition accuracy  improvement of 43.12\% on images more than 10\% covered by dynamic objects. As more of the image is occupied by dynamic objects, the accuracy improvements increase: for images with more than 20\% dynamic object coverage, accuracy improves by 55.75\%, and if coverage is more than 30\%, the improvement increases to 62.22\%. 

\begin{figure*}[th]
\centering
   \includegraphics[width=1\textwidth]{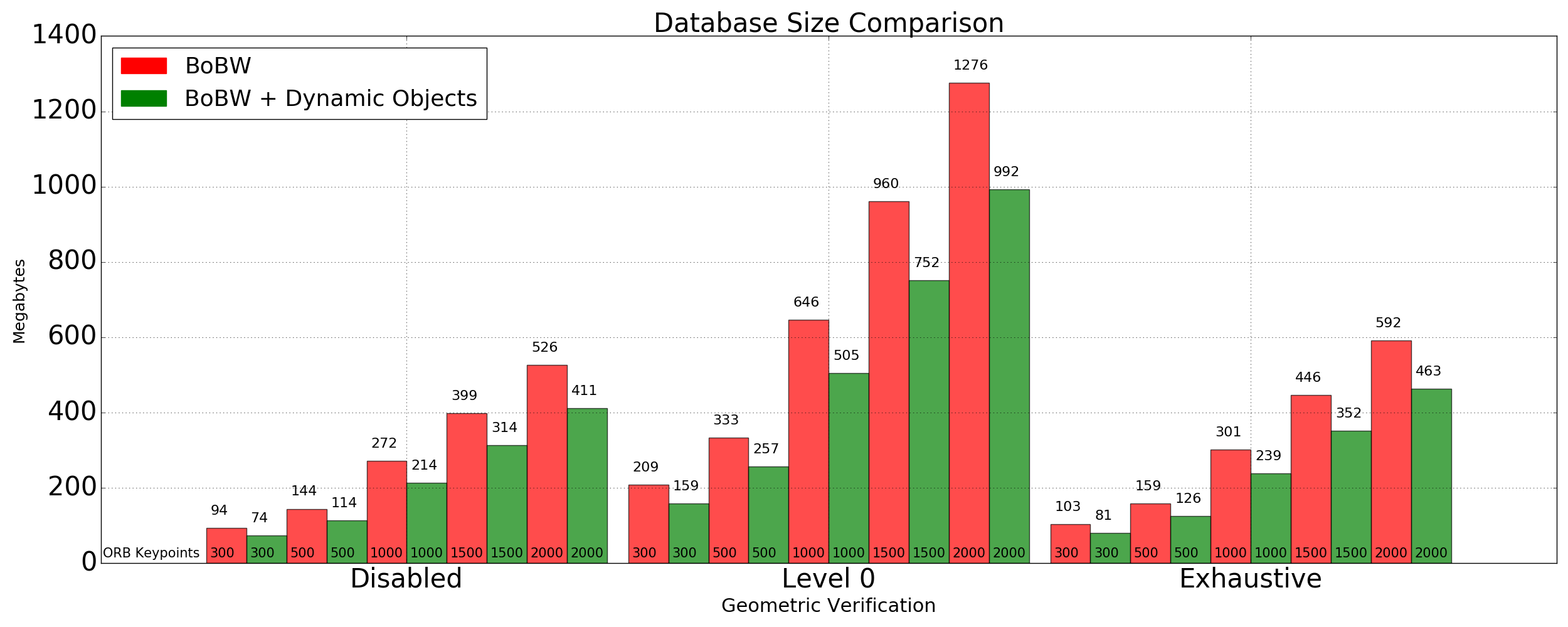}
   \caption{Comparison of databases generated using the Synthia dataset. The proposed approach significantly reduces the size of the database, and  produces better recognition results than the version that uses the original place representation.}~\label{fig:dbCompSynthia}
\end{figure*}

Figure \ref{fig:dbCompSynthia} shows a comparison of the databases generated after processing the \emph{Synthia} dataset. The proposed approach generates much smaller databases for all configurations. For instance, setting the number of maximum \emph{ORB} keypoints to 300 and disabling geometric verification (see Section \ref{BoW}), our approach reduces the database size from 94.36 MB to 74.44 MB (21.1\%).  When the geometric verification uses level 0 of the vocabulary tree, the database size is reduced from 209 MB to 159 MB (23.9\%). In the case of exhaustive geometric verification, and using 300 keypoints, our approach reduces the size of the database from 103 MB to 81 MB. Another example is the configuration that uses a maximum of 1500 \emph{ORB} keypoints and no geometric verification. Here the reduction is 21\% from the original size, saving 84.5 MB of storage space.

\begin{figure*}[th]
\centering
   \includegraphics[width=1\textwidth]{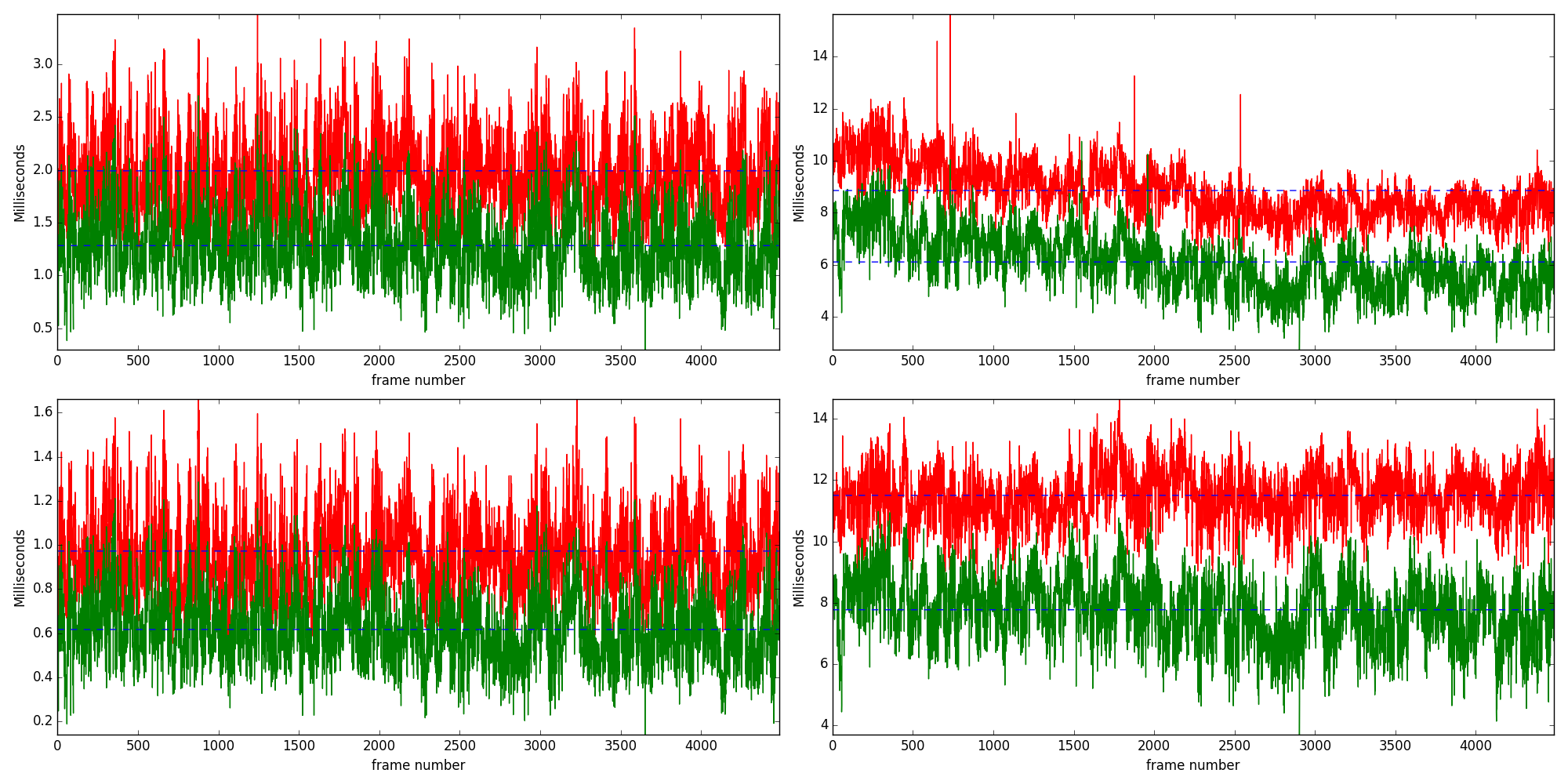}
   \caption{Place recognition latency using 4500 frames of the Synthia dataset. Red lines correspond to the original BoBW approach; green lines correspond to our BoBW+DO approach. The graphs on the left correspond to the configuration that uses $\approx{300}$ ORB features, while the graphs on the the right correspond to the configuration that uses $\approx{1500}$ ORB features. The top row gives results with geometric verification disabled; the bottom row shows the results using exhaustive geometric verification. Dashed lines represent the average time for each method.}~\label{fig:timeCompSynthia}
\end{figure*}

Reducing the size of place representations confers an additional benefit on the time required to find matches in the database. Figure \ref{fig:timeCompSynthia} shows a comparison of the time required to match places by the original approach (BoBW) and our extension (BoBW + DO). Our approach decreases this required time by several milliseconds depending on the selected configuration. For instance, when using 1500 ORB features, our approach decreases the average required time for attempting to recognize a place without geometric verification, from $\approx{9}$ milliseconds to $\approx{6}$ milliseconds. However, our approach requires the costliest object detection step. The stage for detecting dynamic objects took an average of 66 milliseconds per image, which includes resizing the image to 416x416 to meet the object detector requirements. The average time to detect objects is expected to decrease to $\approx{22}$ milliseconds per image when no image resizing is needed.

\begin{figure}
\centering
   \includegraphics[width=0.6\columnwidth]{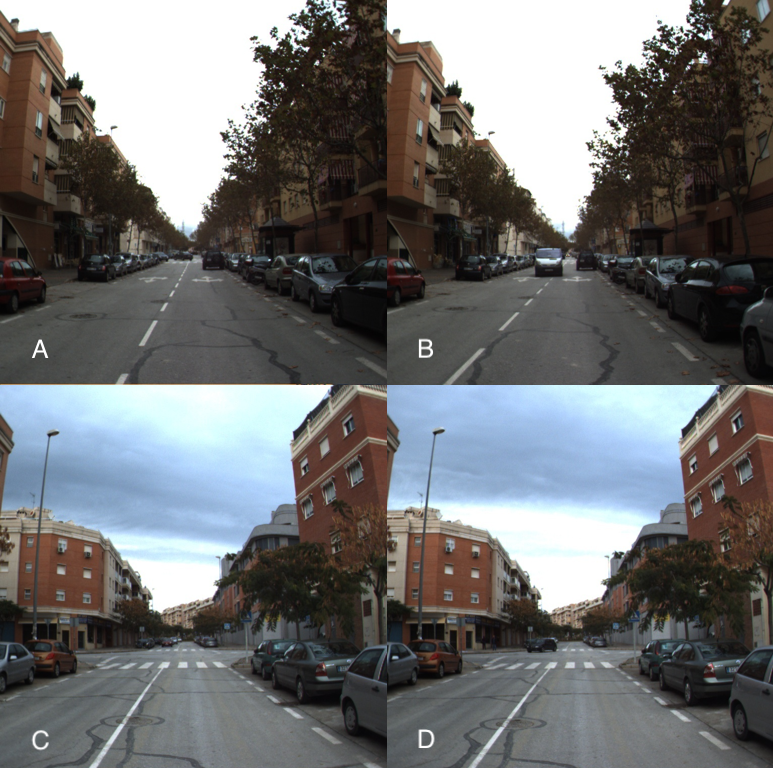}
   \caption{Dynamic objects behaving as static objects in the M\'{a}laga dataset. A-B and C-D: Several of the dynamic objects detected during the first visit, e.g., cars parked on the street, remain in the same place until the next visit of the agent, behaving as static objects.}~\label{fig:malagaRevisit}
\end{figure}

\begin{figure}[!th]
\centering
  \includegraphics[width=0.8\columnwidth]{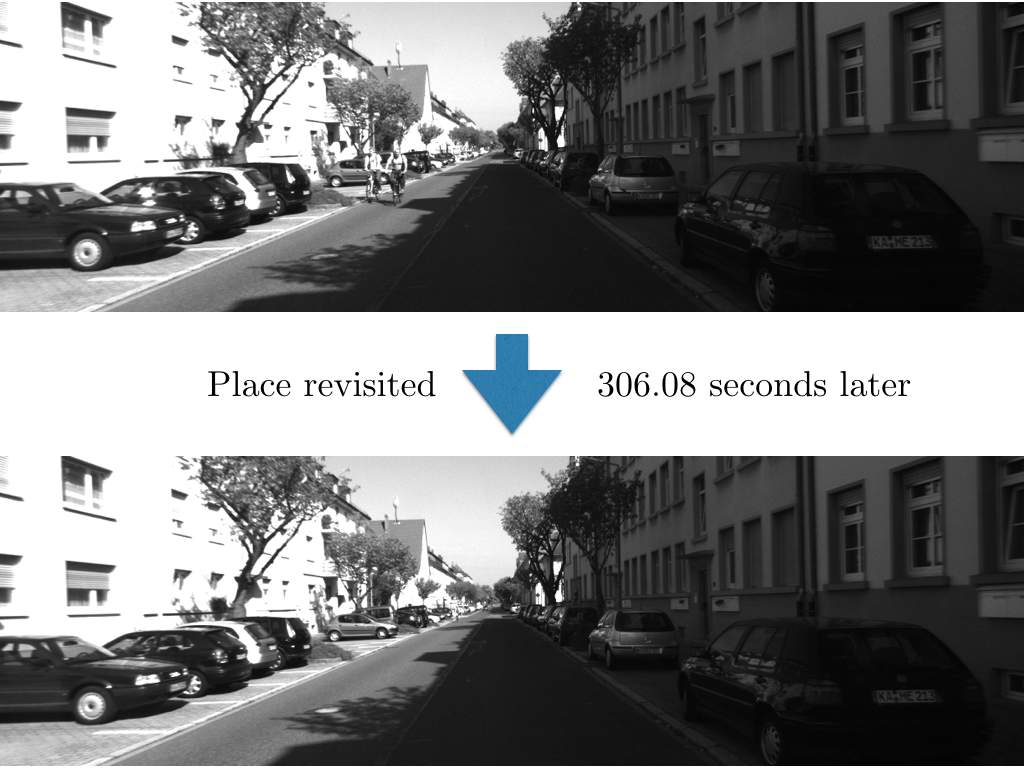}
  \caption{Example from the Kitti dataset \cite{Geiger2013} of dynamic objects behaving as static objects. The agent revisits this place a few minutes later; most of the cars parked on the street are in the same exact place. Our approach may not be suitable for applications in which this situation is expected to arise frequently.}~\label{fig:kittiRevisit}
\end{figure}

\subsection{Real-World Dataset Insights}

The M\'{a}laga urban dataset provides additional insights into the behavior of the proposed approach. In this dataset, all 17,300 frames were captured in a short period, a little more than 14 minutes. Some segments of the route used for our evaluation were revisited by the agent with an inter-visit interval of just a few seconds. With such a short timespan between visits, many dynamic objects remained in the same place, thus behaving more like static objects. For instance, most of the cars that appeared parked in the first visit were also spotted in the following visits as illustrated in Figure \ref{fig:malagaRevisit}. This characteristic is not unique to the M\'{a}laga urban dataset. Other subsets of popular datasets, e.g, Kitti, present similar characteristics as illustrated in Figure \ref{fig:kittiRevisit}, in which a place that is revisited after 306.08 seconds (about 5 minutes), it encounters nominally dynamic objects that have not moved at all. Our approach is expected to thrive when the agent is exploring a highly dynamic environment, or when enough time has passed to allow for dynamic objects to behave as such.

\begin{figure*}[!th]
\centering
   \includegraphics[width=1\textwidth]{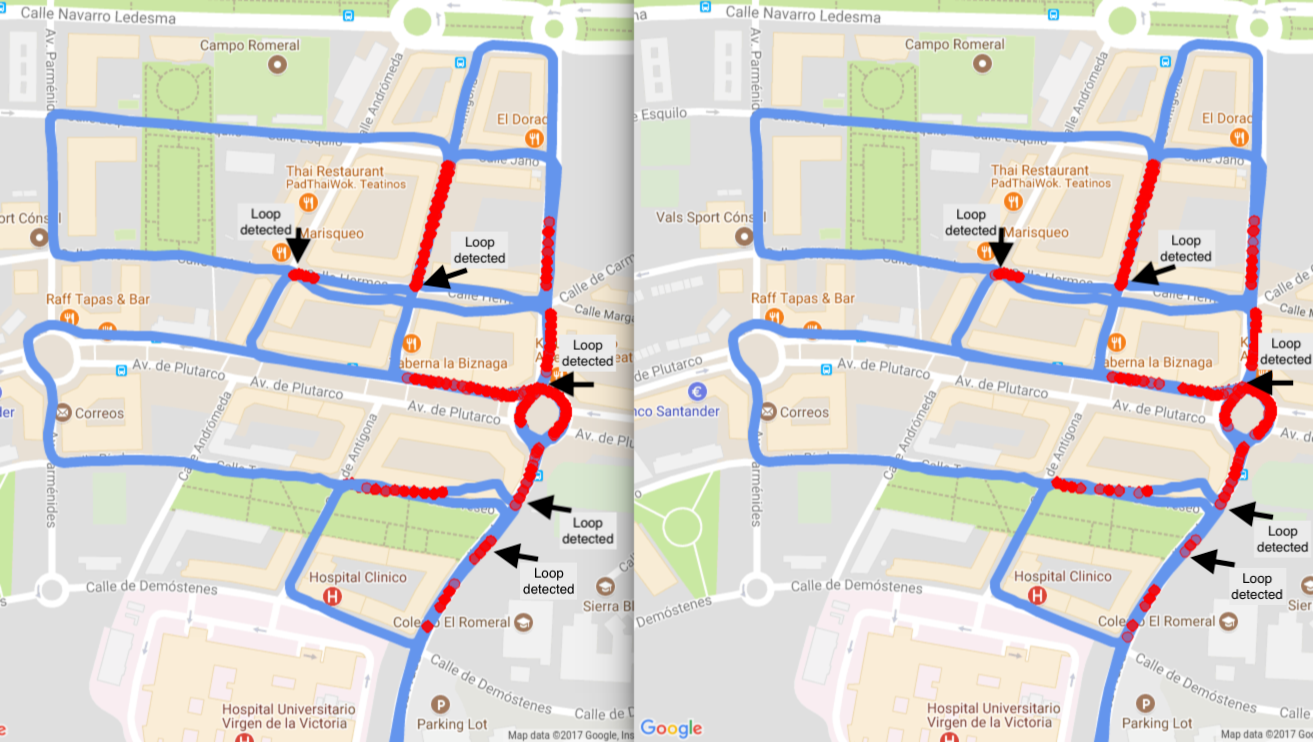}
   \caption{Comparison of place recognition matches found by the original (left) and proposed (right) approaches. The extended approach detected the same loop closures as the original algorithm. The path traversed by the vehicle is in blue, while the places that have been correctly recognized when revisited are in red. Each loop closure in the subset of the M\'{a}laga dataset is indicated with an arrow. }~\label{fig:malagaLoopsMap}
\end{figure*}

Despite the fact that the agent revisited some places in the M\'{a}laga dataset in a very short time, thereby reducing the benefits of our approach, we were able to detect the same number of loop closures as the original BoBW approach. Subset \#10 of the M\'{a}laga dataset contains five loops; all of the closures of these loops were correctly detected. This is illustrated in Figure \ref{fig:malagaLoopsMap}. While our approach does not lose accuracy in less dynamic environments, the additional computational costs incurred by object recognition may not yield a corresponding benefit. However, BoBW+DO still produces a significantly smaller database while exploring this subset of the M\'{a}laga dataset, while maintaining similar recognition results. For instance, setting the number of maximum \emph{ORB} keypoints to 1500 and enabling exhaustive geometric verification, gives a 14.3\% reduction in database size from the original BoBW, from 1705 MB to 1462 MB; when geometric verification uses level 0 o, the database size is reduced by 14.1\%, from 3687 MB to 3166 MB.

\begin{figure*}[!th]
\centering
   \includegraphics[width=1\textwidth]{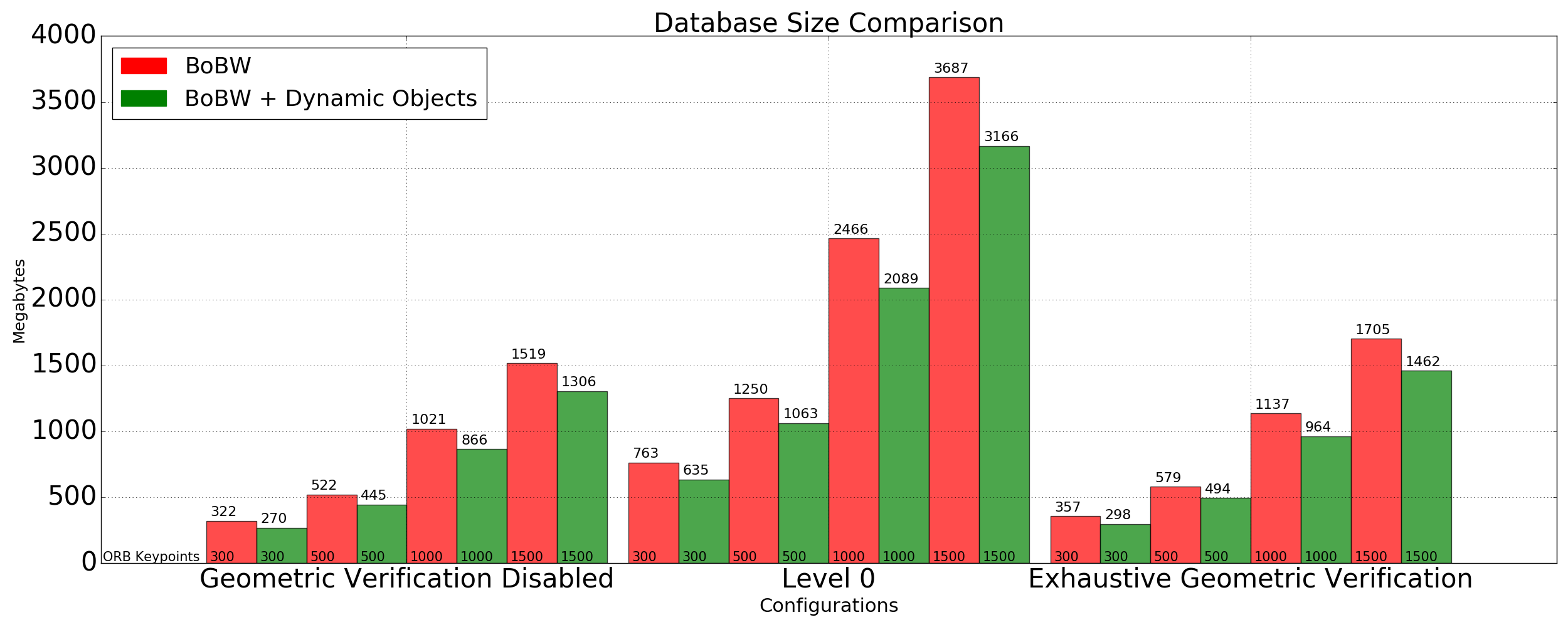}
   \caption{Comparison of databases generated using the M\'{a}laga dataset. \emph{BoBW-Dynamic Objects} performs as well as \emph{BoBW} by recognizing the same number of loop closures, but with the additional benefit of reducing the size of the database.}~\label{fig:dbCompMalaga}
\end{figure*}

\section{Conclusions and Future Work}~\label{discussion} 

Appearance-based place recognition approaches are still plagued by several challenges that are rooted in the complexity of the real world and the limitations of visual sensors. One of those challenges is the intermittent presence of dynamic objects. In this article, we have presented an approach to reduce the negative impact of dynamic objects on place representation and recognition. 

Our approach offers several benefits, including the reduction of storage requirements while improving recognition accuracy. This approach can be used to improve the performance of suitable existing place recognition algorithms in environments with significant numbers of dynamic objects.

Our approach relies on modifying ``traditional'' place recognition algorithms; only those with ``flexible'' representations, which allow us to manipulate them to incorporate object information, are suitable for our modifications. We illustrate the performance improvements of our approach by augmenting the state-of-the-art \emph{Bags of Binary Words} algorithm \cite{Galvez-Lopez2012}. In the future, we anticipate applying our approach to other suitable algorithms to further substantiate the significance of this approach.

Modifying place representations based on the presence of dynamic objects in the observations may not generalize well to applications in which an agent will revisit the environment in a very short amount of time, primarily because most of the dynamic objects may have not moved since the previous visit, e.g., cars parked on the street. Figure \ref{fig:malagaRevisit} from the M\'{a}laga dataset and Figure \ref{fig:kittiRevisit} from the Kitti dataset illustrate these kinds of situations. 

Future work will also explore improvements in the approximation of the area covered by the detected dynamic objects maintaining the requirement of running in real-time. This improvement will result in a more precise identification of the proportion of the extent of the descriptor that is affected by dynamic objects and in further improvement to the resultant place representation.

Finally, we expect that information about dynamic objects could have additional applications. For example, this information could allow navigation modules to plan paths that avoid areas where there is a tendency toward a high presence of dynamic objects. The information about dynamic objects could also be used to determine the kind of place that an agent is visiting, which could also enrich  navigation applications. 


%

  
\bibliographystyle{IEEEtran}
\bibliography{dynamicbiblio}

%
\ifdefined\ADDAFFILIATION
\begin{IEEEbiography}[{\includegraphics[width=1in,height=1.25in,clip,keepaspectratio]{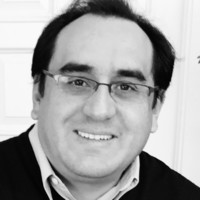}}]{J. Pablo Muñoz}
graduated magna cum laude from Brooklyn College with a B.A. in Philosophy and membership in Phi Beta Kappa. He then went on to earn an M.S. in Computer Science from the Grove School of Engineering at the City College of New York and a Ph.D. in Computer Science from the City University of New York. He is currently a Research Scientist at Intel Labs. His research includes the design and development of frameworks for large-scale video analytics. Pablo is also a contributor to AutoML and Immersive Media projects. Previously, he successfully led the development of localization systems for assisting visually impaired people to navigate indoors and designed and implemented an award-winning prototype to combat the spread of the Zika virus using state-of-the-art computer vision techniques and citizen science.
\end{IEEEbiography}

\begin{IEEEbiography}[{\includegraphics[width=1in,height=1.25in,clip,keepaspectratio]{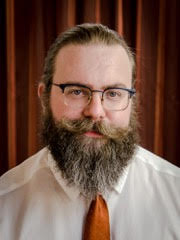}}]{Scott Dexter} was born in New Jersey (US) in 1972 and was substantially raised in Michigan. He earned the BS in Mathematics and Computer Science from Denison University, Granville, Ohio, and the MS and PhD in Computer Science and Engineering from the University in Michigan, Ann Arbor.

He taught at Brooklyn College of the City University of New York from 1998 to 2019, reaching the rank of Professor of Computer and Information Science, and currently serves as Professor of Computer Science at Alma College, Alma, Michigan.

He has been inducted into Phi Beta Kappa, Sigma Xi, Upsilon Pi Upsilon, and Pi Mu Epsilon, and is currently a member of the Association for Computing Machinery and the Free Software Foundation.
\end{IEEEbiography}

\fi 






\end{document}